\documentclass{article}

\usepackage{float}
\usepackage{wrapfig}
\usepackage[preprint]{corl_2026}
\usepackage{amsmath,amssymb,mathtools}
\usepackage{booktabs}
\usepackage{array}
\usepackage{tabularx}
\usepackage{graphicx}
\usepackage{microtype}
\usepackage{eso-pic}
\usepackage{fontawesome5}

\definecolor{lampred}{rgb}{0.808,0.204,0.255}
\definecolor{lampcoral}{rgb}{0.929,0.443,0.416}
\definecolor{lampyellow}{rgb}{0.992,0.863,0.427}
\definecolor{lampgold}{rgb}{0.745,0.592,0.098}
\newcommand{\lampword}{\textcolor{lampred}{L}\textcolor{lampcoral}{A}\textcolor{lampyellow}{M}\textcolor{lampgold}{P}}

\title{\lampword{}: Latent Motion Prior-Guided Real-World Learning for Dexterous Hand Manipulation}

\author{
\textbf{Xinye Yang}$^{1,4,5,*}$ \quad
\textbf{Zhiyuan Ma}$^{2,4,*}$ \quad
\textbf{Hongze Yu}$^{4,\dagger}$ \quad
\textbf{Yuanpei Chen}$^{4}$ \\
\textbf{Yaodong Yang}$^{3}$ \quad
\textbf{Xiaojie Chai}$^{4}$ \quad
\textbf{Xinlei Chen}$^{2}$ \quad
\textbf{Chao Yu}$^{2,\dagger}$ \\[0.35em]
{\normalfont $^{1}$Fudan University, $^{2}$Tsinghua University, $^{3}$Peking University} \\
{\normalfont $^{4}$PsiBot, $^{5}$Zhongguancun Academy} \\
{\normalfont $^{*}$Equal contribution, $^{\dagger}$Corresponding authors}
}

\newcommand{\method}{LAMP}
\newcommand{\obs}{o}
\newcommand{\act}{a}
\newcommand{\arm}{\mathrm{arm}}
\newcommand{\hand}{\mathrm{hand}}
\newcommand{\hist}{H}
\newcommand{\latent}{z}
\newcommand{\real}{\mathbb{R}}
\newcommand{\Aarm}{\mathcal{A}_{\arm}}
\newcommand{\Ahand}{\mathcal{A}_{\hand}}
\newcommand{\Zspace}{\mathcal{Z}}
\newcommand{\Ddemo}{\mathcal{D}_{\mathrm{demo}}}
\newcommand{\Donline}{\mathcal{D}_{\mathrm{online}}}
\newcommand{\prior}{LMPM}
\newcommand{\figbox}[2][0.16\textheight]{%
  \fbox{\begin{minipage}[c][#1][c]{0.96\linewidth}
  \centering\small #2
  \end{minipage}}}
\newcommand{\placepsibotlogo}{%
  \AddToShipoutPictureFG*{%
    \AtPageUpperLeft{%
      \hspace{0.75in}\raisebox{-1.02in}{%
        \includegraphics[width=1.15in,trim=45 285 45 285,clip]{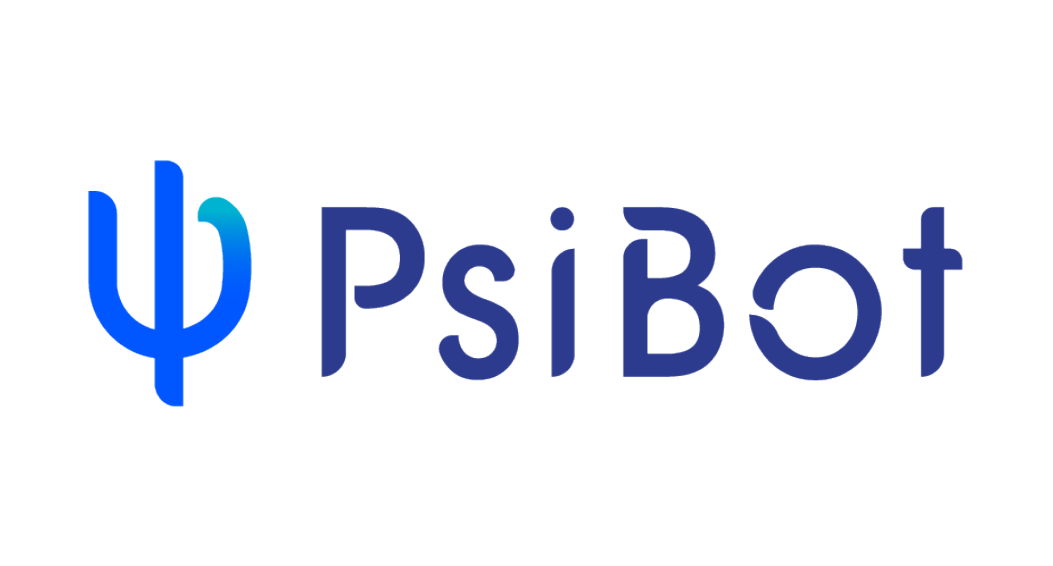}%
        \hspace{0.08in}%
        \raisebox{0.16in}{%
          \includegraphics[width=0.58in,trim=135 80 135 80,clip]{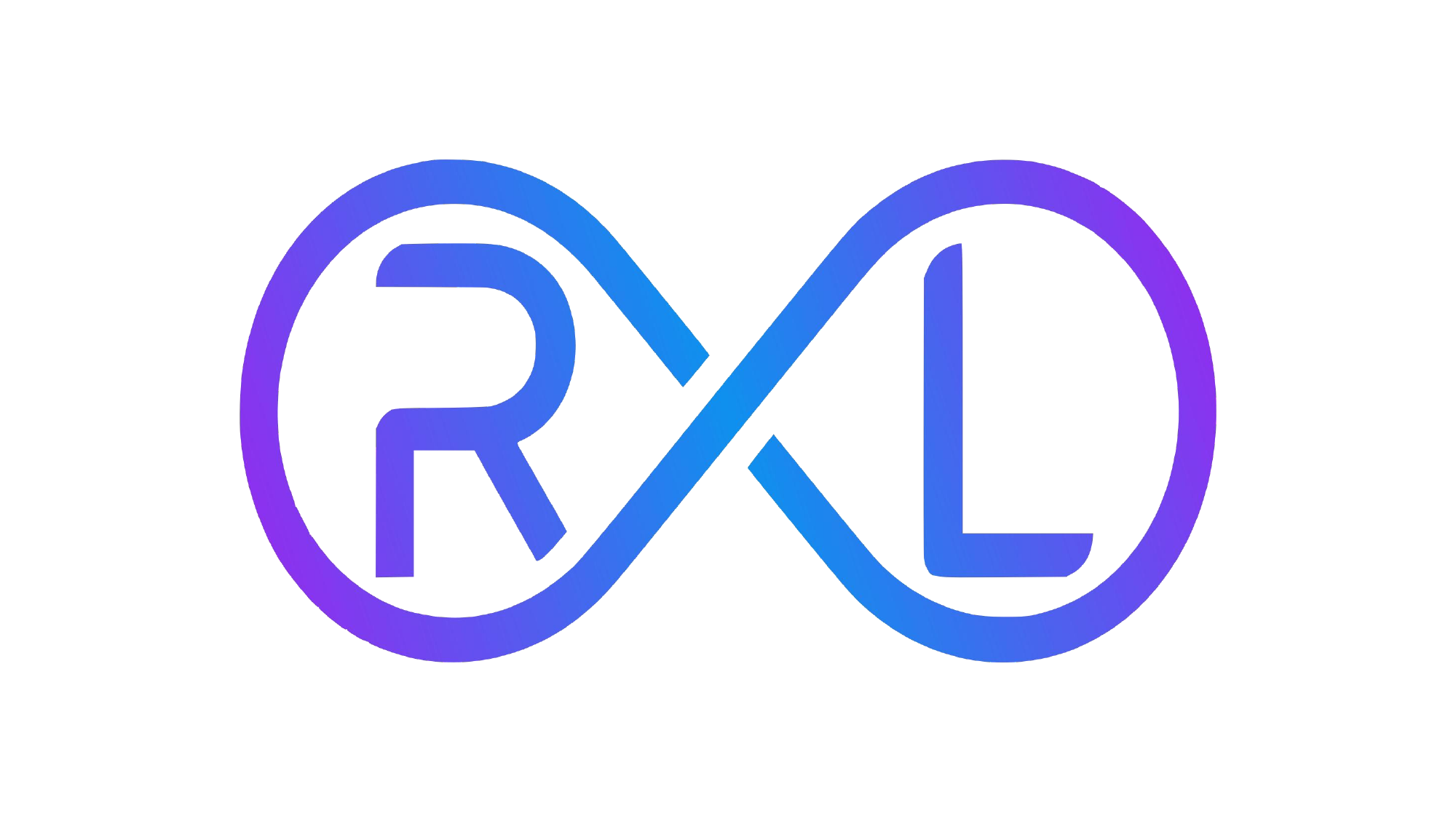}%
        }%
      }%
    }%
  }%
}

\hypersetup{
  pdftitle={LAMP: Latent Motion Prior-Guided Real-World Learning for Dexterous Hand Manipulation},
  pdfauthor={Xinye Yang, Zhiyuan Ma, Hongze Yu, Yuanpei Chen, Yaodong Yang, Xiaojie Chai, Xinlei Chen, Chao Yu},
  pdfsubject={arXiv preprint},
  pdfkeywords={Dexterous Manipulation, Motion Prior, Real-World Robot Learning}
}

\begin{document}
\raggedbottom
\placepsibotlogo
\maketitle

\vspace{-0.20in}
{\centering\small
\href{https://dex-lamp.github.io/}{\faIcon{globe}\hspace{0.35ex}Project Page}\qquad
\href{https://github.com/dex-lamp/LAMP}{\faGithub\hspace{0.35ex}Code}\par}
\vspace{0.03in}

\begin{figure}[H]
    \centering
    \includegraphics[width=0.95\linewidth]{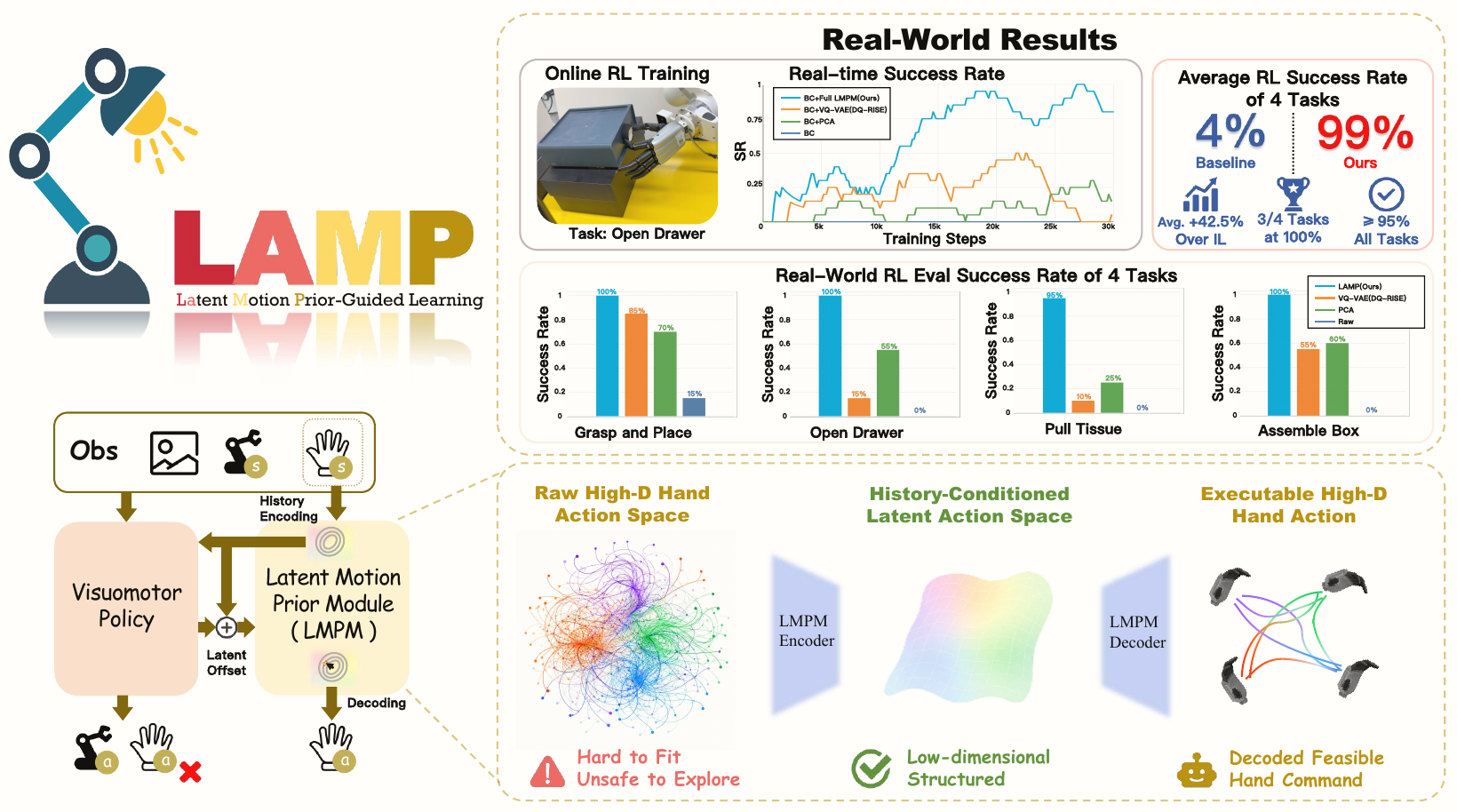}
    \caption{\method{} uses a learned motion prior to expose high-dimensional dexterous hand motion through a compact latent action space, where imitation learning predicts latent offsets and residual reinforcement learning adds latent residuals before decoding executable hand actions.}
    \label{fig:abstract}
\end{figure}

\begin{abstract}
Real-world learning for dexterous hands remains brittle because high-dimensional hand actions amplify imitation errors and make reinforcement-learning exploration prone to contact-breaking motion. While combining imitation learning (IL) with online reinforcement learning (RL) can reduce manual supervision, unconstrained exploration in raw hand-action spaces is sample-inefficient and risky for physical hardware. We introduce a latent motion prior module (\prior{}) that maps recent hand-action histories to a compact, history-conditioned latent prior and decodes continuous latent commands into executable high-dimensional hand targets. Built on this prior, \method{} is a three-stage real-world dexterous learning framework: it pretrains \prior{} from demonstrations, trains a visuomotor policy that predicts native arm commands and latent hand-action offsets, and improves the policy with online residual RL in the same latent hand-action space. This shared, decodable interface lets residual exploration make local corrections near demonstrated, contact-consistent hand motions rather than perturbing every finger joint independently. We evaluate \method{} on four real-robot dexterous manipulation tasks against raw, linear, and discrete hand-action interfaces. Starting from small task-specific demonstration sets, \method{} achieves a 56.25\% average IL success rate and raises it to 98.75\% after online RL, reaching 100\% final success on three tasks and 95\% on the remaining task.
\end{abstract}
\vspace{-1.5ex}

\section{Introduction}
\label{sec:intro}

Dexterous robot hands promise contact-rich manipulation beyond the capabilities of simple grippers. Their high-dimensional actuation lets them conform to diverse object geometries, adjust grasp postures, and maintain multi-finger contact with articulated or compliant objects. To deploy these capabilities on hardware, current research mainly relies on sim-to-real transfer and imitation learning to bypass the challenges of analytical modeling.

However, both paradigms face severe bottlenecks when scaled to dexterous manipulation. For sim-to-real transfer, the chaotic, discontinuous nature of multi-finger contact dynamics, together with micro-friction and soft-material compliance, makes high-fidelity simulation notoriously difficult, leading to a pronounced reality gap that degrades policy performance upon physical deployment~\citep{lin2025simtorealreinforcementlearningvisionbased}. For imitation learning, standard visuomotor policies~\citep{zhao2023act,chi2023diffusion,ze2024dp3} typically excel at replicating demonstrated trajectories but fundamentally lack the capacity for active exploration and error correction. Because end-to-end behavioral cloning does not model alternative recovery behaviors, minor execution errors in high-DoF action spaces quickly compound, causing the hand to drift into out-of-distribution (OOD) configurations. Without the exploratory optimization inherent in reinforcement learning, pure imitation learning cannot reason about contact recovery or adapt to dynamic perturbations from limited human demonstrations~\citep{feng2026learningdexterousmanipulationquantized}.

To overcome these limitations, online reinforcement learning directly in the real world has emerged as a compelling alternative for fine-tuning behaviors. However, standard RL relies on unconstrained, stochastic exploration to establish reliable action-value estimates. In unconstrained high-DoF spaces, unguided exploration inevitably triggers out-of-domain movements that perturb object poses or break delicate contact patterns. Once the object tilts or falls, mid-episode recovery is practically impossible for a physical hand, leading to irreversible state transitions, extremely sparse successful trajectories, and prohibitively low sample efficiency. This erratic and random exploration contrasts sharply with biological learning. Developmental psychology~\citep{von2004action,thelen1995motor} reveals that human infants do not explore via unstructured, random joint movements; instead, their early actions, known as ``structured motor babbling'', are inherently guided by intrinsic neuromuscular priors.

A natural remedy to this exploration crisis lies in exploiting the intrinsic low-dimensional structure of dexterous hand movements. Prior studies suggest that hand motion often contains low-dimensional structure: multi-joint grasping postures can be explained by a small number of modes~\citep{santello1998postural,ciocarlie2009hand}. This observation has motivated compressed and latent action spaces, from linear hand subspaces to learned continuous or discrete latent policies~\citep{zhou2020plas,allshire2021laser,lee2024behaviorgenerationlatentactions,feng2026learningdexterousmanipulationquantized}. Yet a dexterous learning pipeline needs more than dimension reduction: the interface must decode to executable hand targets, support locally smooth residual updates around contact, and use recent motion history so exploration stays near contact-consistent hand motions. This motivates a continuous, decodable, history-conditioned latent action space shared by supervised policy learning and online residual exploration.

To instantiate this interface, we introduce the latent motion prior model (\prior{}), which maps recent hand-action history to a distribution in a compact latent action space and decodes sampled latent variables back to raw hand targets. We then build \method{}, a three-stage real-world learning framework around the same prior interface: Stage 1 pretrains \prior{} from offline hand-motion data; Stage 2 behavior-clones arm commands and predicts latent hand-action offsets around the LMPM prior; and Stage 3 applies residual reinforcement learning in the same latent action space.

Our contributions are:
\begin{itemize}
    \setlength{\topsep}{2pt}
    \setlength{\partopsep}{0pt}
    \setlength{\itemsep}{2pt}
    \setlength{\parsep}{0pt}
    \item We introduce \prior{}, a history-conditioned latent motion prior that learns a continuous, decodable manifold of dexterous hand motion from offline trajectories.
    \item We develop \method{}, a real-world learning framework that unifies imitation learning and residual reinforcement learning through the shared LMPM latent hand-action interface.
    \item We validate \method{} on four real-world dexterous tasks, achieving 100\% final success on three tasks and 95\% on the remaining task, demonstrating stronger real-world learning performance than alternative hand-action interfaces.

\end{itemize}

\section{Related Work}
\label{sec:related}

\paragraph{Hand action-space representations.}
Low-dimensional hand synergies have long been observed in human grasping and tool use~\citep{santello1998postural} and have been used to reduce robotic grasp search~\citep{ciocarlie2009hand}. Recent work extends this idea to learned cross-hand pose synergies~\citep{pchang2025pchands}, continuous latent action spaces for offline or online policy search~\citep{zhou2020plas,allshire2021laser}, reusable skill priors~\citep{pertsch2020acceleratingreinforcementlearninglearned,lynch2019play}, and shared action spaces from unlabeled or cross-embodiment data~\citep{liang2025clam,bauer2025latent}. \method{} builds on this continuous representation view, but uses a history-conditioned, decodable hand-action interface as the common control space for both behavior-cloning offsets and online residual-RL corrections on real dexterous tasks.

Discrete representations instead tokenize actions for multimodal behavior modeling and sequence prediction. Examples include vector-quantized behavior policies~\citep{lee2024behaviorgenerationlatentactions}, tokenized or disentangled robot action spaces~\citep{wu2024discrete}, efficient VLA action tokenization~\citep{pertsch2025fast}, and dexterous policies with discrete action chunks or quantized hand commands~\citep{yang2024vqace,feng2026learningdexterousmanipulationquantized,jiang2026crosshandlatentrepresentationvisionlanguageaction}. These methods are effective for action chunking and large-scale sequence modeling, but codebook switches can turn small latent changes into abrupt hand-command jumps during contact-rich residual RL. \method{} therefore uses a continuous hand interface constrained by the demonstrated motion manifold.

\paragraph{Real-world imitation and reinforcement learning.}
Modern real-world imitation learning includes action-chunking transformers~\citep{zhao2023act}, diffusion-based visuomotor policies~\citep{chi2023diffusion}, 3D-conditioned diffusion policies~\citep{ze2024dp3}, and large-scale generalist robot policies~\citep{brohan2023rt1,brohan2023rt2,kim2024openvla,octo2024}. These approaches differ in architecture and data scale, but typically predict environment actions or action chunks directly. Real-world RL systems combine demonstrations, online replay, sparse visual rewards, asynchronous actor-learner infrastructure, and human feedback for data-efficient policy improvement~\citep{ball2023rlpd,luo2024serl,luo2024hilserl,zang2026rlinfuser,zang2025rlinfvla,yu2026zprl}. \method{} changes the action space in which imitation and residual exploration occur within this real-world learning pipeline.

\section{Method}
\label{sec:method}

\begin{figure}[t]
    \centering
    \includegraphics[width=0.9\linewidth]{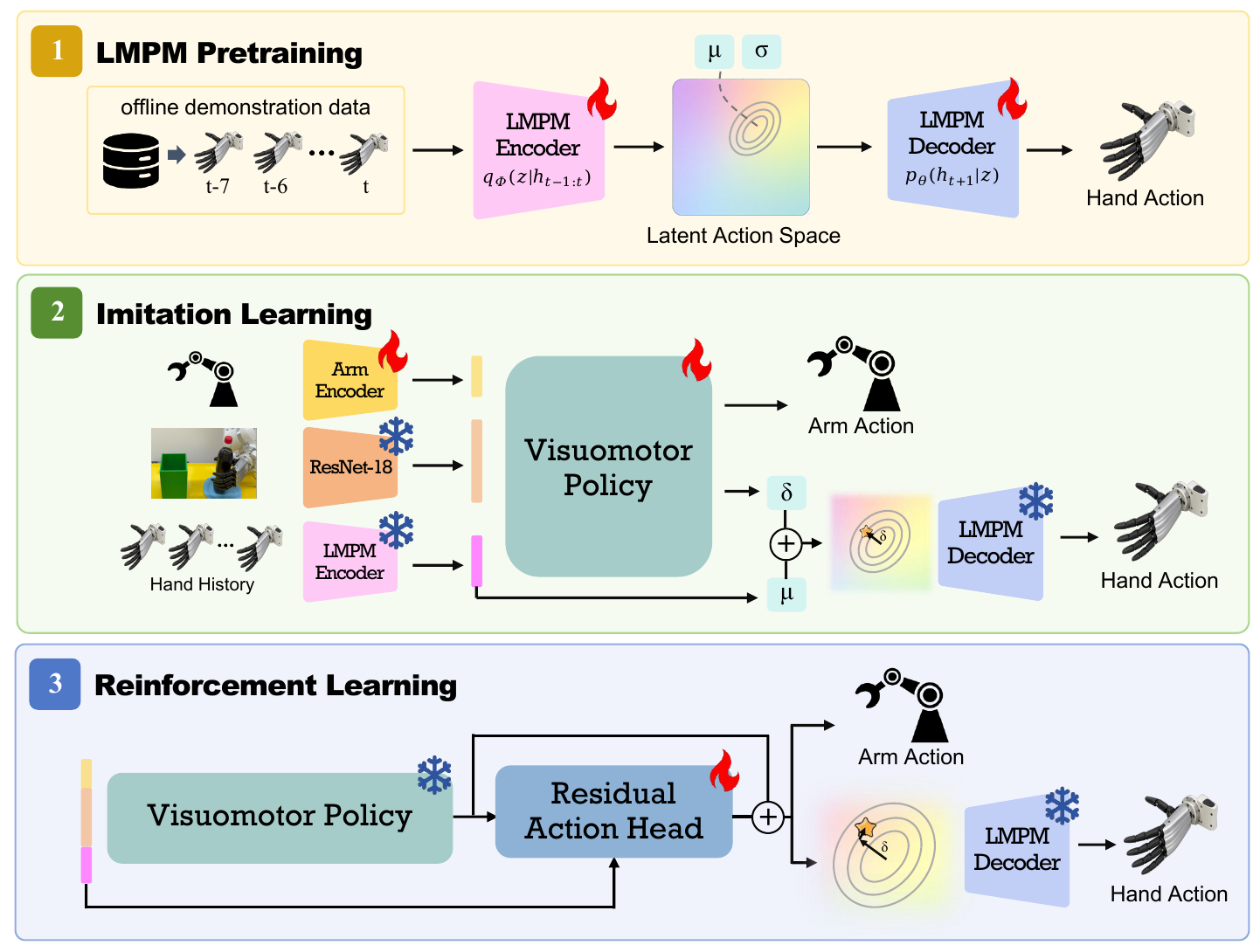}
    \caption{Overview of \method{}. The framework first learns \prior{} from demonstrated hand motion and then uses the frozen prior interface in both behavior cloning and real-world residual RL.}
    \label{fig:pipeline}
\end{figure}

\method{} uses \prior{} to expose high-dimensional hand control through a compact, history-conditioned latent interface shared by imitation learning and real-world residual RL. In this section, we first learn \prior{} as an encoder from hand-target histories to a latent prior and a decoder from latent vectors to executable hand commands (Section~\ref{subsec:lmpm}). We then use this frozen prior to train an IL policy that predicts arm commands in the native arm space and hand commands as vision-guided latent offsets around the prior center (Section~\ref{subsec:il}). Online residual RL also operates in the latent coordinates, adding native arm residuals and latent hand residuals before decoding the final hand target to avoid contact-breaking perturbations in the original hand-command space (Section~\ref{subsec:rl}). Figure~\ref{fig:pipeline} summarizes the three-stage pipeline.

\subsection{Stage 1: Learning the Latent Motion Prior}  
\label{subsec:lmpm}

Stage 1 builds the hand-action interface from the hand-motion component of teleoperated trajectories. Let $h_{t+1}$ denote the target hand command at the next control step, and let
\begin{equation}
    \hist_t = (h_{t-K+1}, \ldots, h_t)
\end{equation}
be the recent hand-target history. The encoder maps $\hist_t$ to a local latent prior, and the decoder maps latent vectors back to executable hand targets:
\begin{equation}
q_\phi(z_t|\hist_t)=\mathcal{N}\!\left(\mu_\phi(\hist_t),\operatorname{diag}(\sigma_\phi^2(\hist_t))\right),
\qquad
\hat{h}_{t+1}=D_\theta(z_t),\quad z_t\sim q_\phi(\cdot|\hist_t).
\end{equation}
We train the encoder and decoder with a reconstruction objective and a KL-regularized bottleneck:
\begin{equation}
\mathcal{L}_{\mathrm{prior}}=
\mathbb{E}_{(\hist_t,h_{t+1})}\!\left[
\|D_\theta(z_t)-h_{t+1}\|_2^2+
\beta D_{\mathrm{KL}}\!\left(q_\phi(z_t|\hist_t)\,\|\,\mathcal{N}(0,I)\right)
\right].
\end{equation}
The reconstruction term preserves executability in the original hand-command space, while the KL term encourages a compact and smooth latent coordinate system. After this stage, $E_\phi$ and $D_\theta$ are fixed. The latent action space supplies shared hand-motion coordinates, and the history-conditioned distribution supplies the local prior center for the current motion phase. Two histories can share the same latent coordinates yet decode around different hand-motion phases through different encoder statistics.

\subsection{Stage 2: Imitation Learning in the Latent Action Space}
\label{subsec:il}

Stage 2 learns the initial visuomotor policy from demonstrations
$\Ddemo=\{(\obs_t,\act_t^{\arm},h_{t+1},\hist_t)\}$.
At time $t$, the observation $\obs_t$ contains RGB images and the arm proprioceptive state $x_t^{\arm}$; hand motion enters separately through the recent hand-target history $\hist_t$ used by the frozen prior. The executed action consists of a native arm command and a hand target,
\begin{equation}
    \act_t = (\act_t^{\arm}, \act_t^{\hand}), \qquad \act_t^{\hand}=h_{t+1}.
\end{equation}
Rather than regressing $h_{t+1}$ directly, the frozen encoder first maps the recent hand history to $(\mu_t,\sigma_t)=E_\phi(\hist_t)$. The behavior cloning policy receives the observation and the prior statistics, and predicts
\begin{equation}
(\hat{\act}_t^{\arm},\Delta\hat{z}_t)=\pi_\psi^{\mathrm{BC}}(\obs_t,\mu_t,\sigma_t),
\qquad
\hat{h}_{t+1}=D_\theta(\mu_t+\Delta\hat{z}_t).
\end{equation}
The supervised objective matches the demonstrated arm command and decoded hand target:
\begin{equation}
\mathcal{L}_{\mathrm{BC}}=
\|\hat{\act}_t^{\arm}-\act_t^{\arm}\|_2^2+
\lambda_h\|\hat{h}_{t+1}-h_{t+1}\|_2^2+
\lambda_z\|\Delta\hat{z}_t\|_2^2 .
\end{equation}
The latent offset lets the policy adapt the hand-motion prior to the current visual observation while keeping decoded commands on the learned hand-motion interface. The arm head remains in the native arm coordinate because the arm must express task-level variation such as approach direction, object pose, and placement geometry.

\subsection{Stage 3: Residual RL in the Latent Action Space}
\label{subsec:rl}

Stage 3 improves the BC policy through online interaction while preserving the same hand interface. New transitions are appended to an online buffer $\Donline$ and mixed with demonstration data during RLPD updates. Starting from the BC policy and the frozen \prior{} decoder, the residual actor observes $s_t=(\obs_t,\hist_t)$ and outputs a residual action $u_t=(\rho_t^{\arm},\rho_t^z)$ in $\Aarm\times\Zspace$, rather than in the full raw hand-joint space. During interaction,
\begin{equation}
(\act_{t,\mathrm{bc}}^{\arm},\Delta z_{t,\mathrm{bc}})=\pi_\psi^{\mathrm{BC}}(\obs_t,\mu_t,\sigma_t),
\quad
z_{t,\mathrm{bc}}=\mu_t+\Delta z_{t,\mathrm{bc}},
\quad
u_t\sim\pi_\eta(\cdot|s_t),
\end{equation}
\begin{equation}
\tilde{\act}_t=
\left(\act_{t,\mathrm{bc}}^{\arm}+s_{\arm}\rho_t^{\arm},\,
D_\theta(z_{t,\mathrm{bc}}+s_z\rho_t^z)\right).
\end{equation}
The executed action is therefore an arm command with a native residual and a decoded hand target with a latent residual. Applying the hand residual in $\Zspace$ expresses exploration as local corrections to a history-conditioned hand command rather than independent perturbations of every finger joint. We update the residual actor and critic with RLPD-style SAC losses on batches $\mathcal{B}\subset \Ddemo\cup\Donline$:
\begin{equation}
\mathcal{L}_Q=\mathbb{E}_{\mathcal{B}}\!\left[(Q_\omega(s_t,u_t)-y_t)^2\right],
\quad
y_t=r_t+\gamma\,\mathbb{E}_{u'\sim\pi_\eta}\!\left[\bar{Q}(s_{t+1},u')-\alpha\log\pi_\eta(u'|s_{t+1})\right],
\end{equation}
\begin{equation}
\mathcal{L}_\pi=\mathbb{E}_{s_t\sim\mathcal{B},\,u_t\sim\pi_\eta}\!\left[\alpha\log\pi_\eta(u_t|s_t)-Q_\omega(s_t,u_t)\right].
\end{equation}
Sparse task rewards are produced by the visual classifier described in Appendix~\ref{app:reward_classifier}.

\section{Experiments}
\label{sec:experiments}

To evaluate whether a learned latent motion prior can make real-world dexterous policy improvement more reliable, we ask three questions: (1) whether the learned hand-action interface improves imitation learning over raw, linear, and discrete alternatives; (2) whether the same interface supports stable and efficient online residual RL on real hardware; and (3) which components of \prior{} are responsible for the gains.

\subsection{Real-World System and Tasks}
\label{subsec:system}

Our setup uses a Franka Research 3 arm with a Ruiyan dexterous hand and two RGB cameras: a front-view RealSense D435 and a wrist-mounted RealSense D405. Both the visual reward classifier and the visuomotor policy receive the same dual-view RGB observations. Demonstrations are collected with human-in-the-loop teleoperation: a Synglove commands finger motion, and a SpaceMouse commands TCP translation and rotation. Additional implementation, reward, and teleoperation details are provided in Appendices~\ref{app:method_details}--\ref{app:data_collection}. During online RL and evaluation, the TCP pose is randomized around a nominal start pose to test spatial generalization.

\begin{figure}[H]
    \centering
    \includegraphics[width=\linewidth]{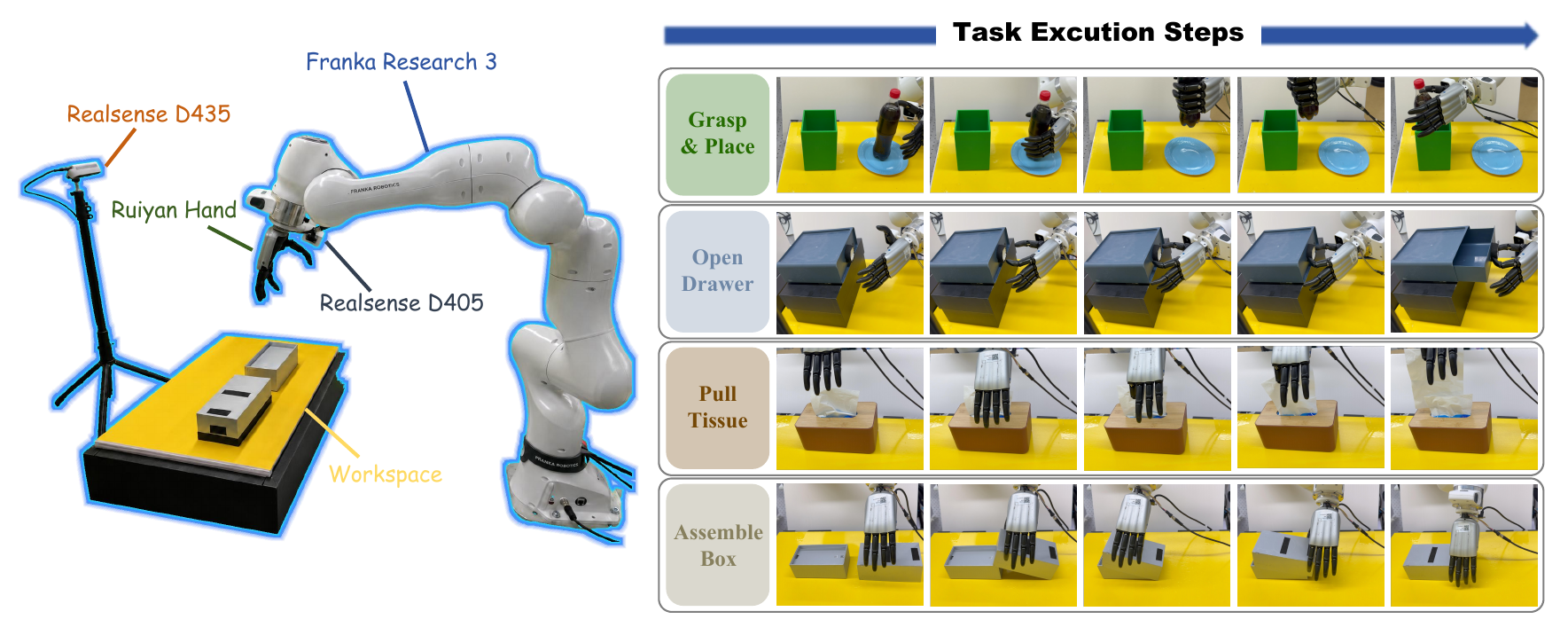}
    \caption{Real-world setup and task scenes. The setup uses a Franka Research 3 arm, a Ruiyan dexterous hand, a front-view RealSense D435 camera, and a wrist-mounted RealSense D405 camera; the right panel shows execution-flow illustrations for the four tasks.}
    \label{fig:tasks}
\end{figure}

We evaluate on four representative real-world tasks (Figure~\ref{fig:tasks}). Each requires sustained hand-object contact during execution, so small finger jitter or abrupt command jumps can break contact, tilt the object, or cause drops. Grasp \& Place requires coordinated arm placement while keeping a bottle upright in a narrow, tall box; Open Drawer requires locating and pinching a small drawer handle; Pull Tissue involves a compliant object with changing geometry; and Assemble Box is a long-horizon insertion task where early contact errors can lead to failure or stalled execution.

\subsection{Comparison of Action Spaces}
\label{subsec:main_results}

We compare four hand-action interfaces under the same imitation-to-residual-RL pipeline. The Raw interface predicts hand commands directly in the original joint space. PCA is a classical linear compression method that reduces the hand-action dimension through a linear transform. VQ-VAE (DQ-RISE) uses the DQ-RISE algorithm~\citep{feng2026learningdexterousmanipulationquantized} to represent hand actions with a discrete codebook. \method{} (Ours) uses the proposed history-conditioned continuous latent action interface. Figure~\ref{fig:network_architectures} summarizes the corresponding action-interface designs. We evaluate these four interfaces under the imitation-learning framework described in Section~\ref{subsec:il} and perform online post-training with the reinforcement-learning algorithm in Section~\ref{subsec:rl}. We use the same offline demonstrations and training hyperparameters for all methods; adapter details are provided in Appendix~\ref{app:method_details}.

\begin{figure}[H]
    \centering
    \includegraphics[width=\linewidth]{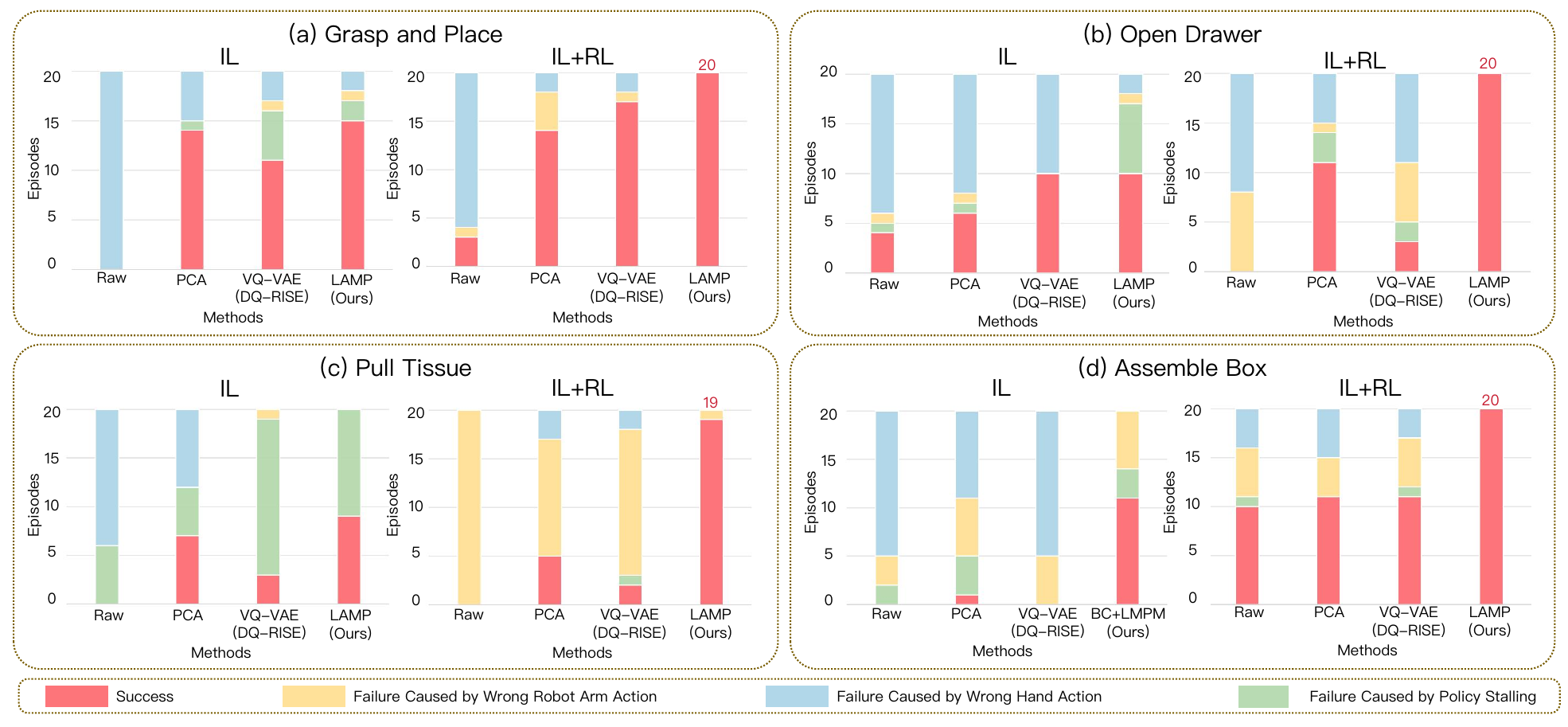}
    \caption{Real-world evaluation on four dexterous tasks. Each bar reports success counts and failures categorized as arm-action error, dexterous-hand error, or stalled execution.}
    \label{fig:results}
\end{figure}

\subsubsection{Imitation Learning Performance}
\label{subsubsec:il_results}

From the results in Figure~\ref{fig:results}, the Raw interface suffers from the high dimensionality of the hand-action regression target: during task execution, the hand motion jitters noticeably, often failing to establish and maintain contact with the object. After PCA linearly compresses the hand action to two dimensions, imitation learning improves substantially over the Raw interface on all four tasks, showing that a compact hand-action representation makes dexterous policy learning easier. VQ-VAE (DQ-RISE) uses a discrete hand-action representation. Although this greatly reduces the action dimension, the hand often switches frequently among nearby actions during execution, causing visible jitter and non-smooth motion. It can perform reasonably well on tasks that require precise finger opposition, but in rigid-object grasping the jitter can easily make the object slip. The remaining gap from PCA to \method{} suggests that compression alone is not sufficient: PCA provides fixed linear hand coordinates, whereas the pretrained history-conditioned encoder in \prior{} predicts a latent prior center from recent hand motion, making hand-action continuity directly available to the visuomotor policy. With this interface, \method{} yields stable and smooth hand motion, reliable object contact, and the highest success rate.

\subsubsection{Reinforcement Learning Performance}
\label{subsubsec:rl_results}

Real-world residual RL is most sample-efficient when exploration remains local to the imitation policy: residuals must improve the behavior without repeatedly breaking contact, so the replay buffer can keep receiving successful or near-successful rollouts. The Raw variant explores in the original hand-action space and starts from a low initial success rate, so RL exploration is extremely inefficient and difficult to converge. PCA reduces the dimension, but its fixed linear projection cannot reparameterize the irregular hand-motion manifold: residuals along retained components can still decode into contact-breaking finger motion. Although VQ-VAE (DQ-RISE) and \method{} substantially improve imitation learning, their nonlinear interfaces may spread many similar hand motions across a broad latent or code region; without task feedback, the imitation policy can remain in a low-progress part of this region and cause stalled executions. During online RL, task rewards provide a directional signal for moving out of these local regions. VQ-VAE can improve with RL, but codebook switches still introduce gesture jumps that limit final evaluation success. By contrast, \method{} uses a continuous decoder and a history-conditioned prior to organize demonstrated hand motions into a better-conditioned latent space, giving residual RL a smoother, more contact-preserving neighborhood for reward-guided refinement. As a result, RL reaches or approaches 100\% success on the tasks. Appendix~\ref{app:qualitative_cases} provides qualitative examples of successful executions and representative failure cases, and Appendix~\ref{app:smoothness} quantifies the command-variation differences behind these contact-stability observations.

\subsection{Action-Interface Ablations}
\label{subsec:action_space}

\begin{wrapfigure}[20]{r}{0.6\linewidth}
    \vspace{-0.5em}
    \centering
    \includegraphics[width=\linewidth]{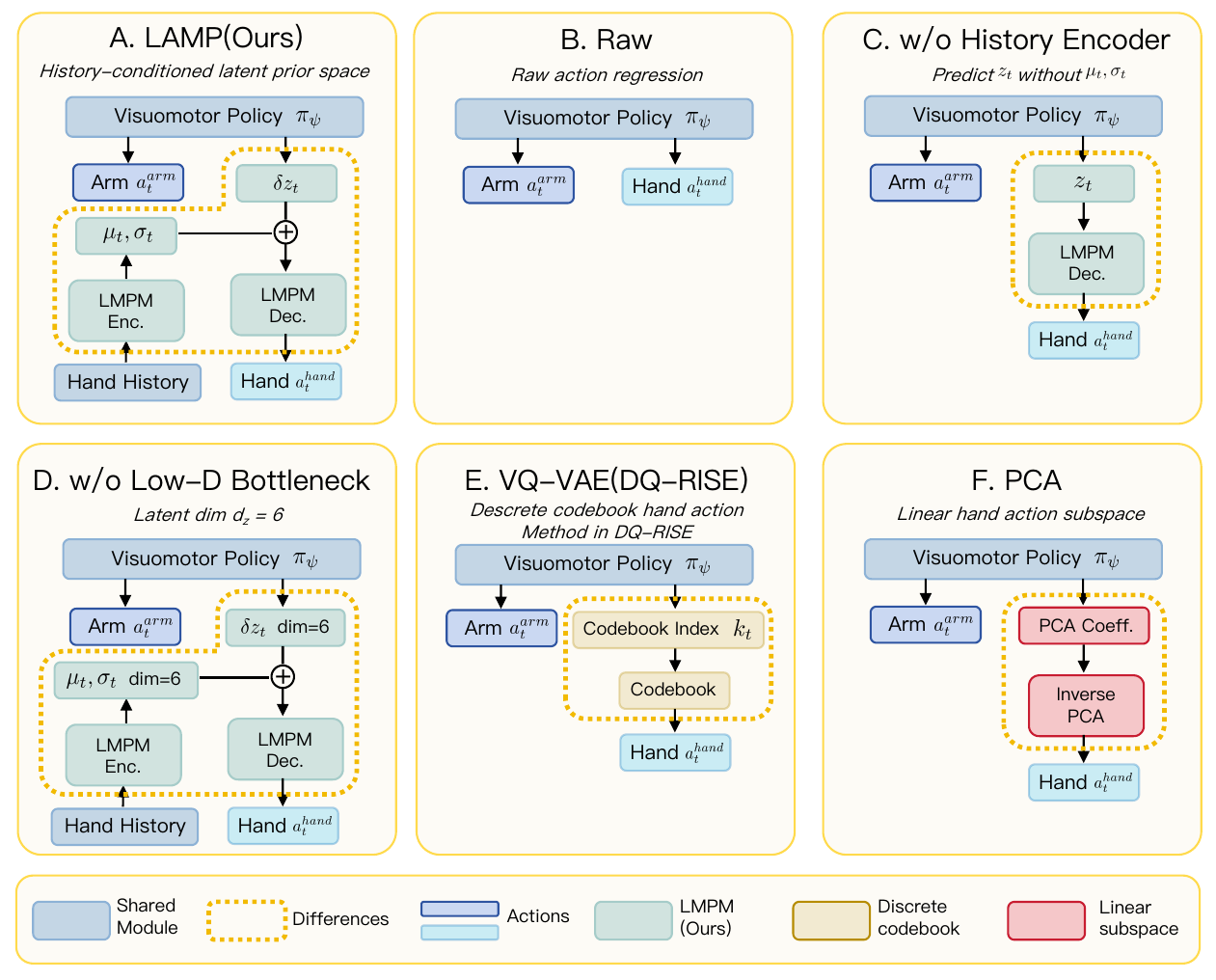}
    \caption{Network architectures for our method, baselines, and ablations.}
    \label{fig:network_architectures}
    \vspace{-0.75em}
\end{wrapfigure}

To isolate which parts of the learned interface matter, we ablate two design choices in the full \prior{} model: the compact latent bottleneck and the history-conditioned prior center. Figure~\ref{fig:network_architectures} summarizes the corresponding architectures together with the baselines, and Table~\ref{tab:ablations} reports the imitation and online RL results, including Raw BC as the raw-space reference.

Both \prior{} design choices matter. Expanding the latent dimension usually lowers final performance, suggesting that a larger hand space still burdens arm-hand coordination and online exploration. Without the history-conditioned encoder, the visual policy predicts latent points without a local prior center, which weakens imitation initialization; residuals are still applied through the decoder and remain tied to the demonstrated motion manifold, so training is more stable than raw hand control. Raw BC performs poorly across tasks. Additional failure-case analysis for the ablations is provided in Appendix~\ref{app:qualitative_cases}.
\WFclear

\begin{table}[H]
\vspace{0.6em}
\centering
\small
\resizebox{\linewidth}{!}{
\begin{tabular}{lcccccccc}
\toprule
Variant & \multicolumn{2}{c}{Grasp \& Place} & \multicolumn{2}{c}{Open Drawer} & \multicolumn{2}{c}{Pull Tissue} & \multicolumn{2}{c}{Assemble Box} \\
\cmidrule(lr){2-3}\cmidrule(lr){4-5}\cmidrule(lr){6-7}\cmidrule(lr){8-9}
 & IL & RL & IL & RL & IL & RL & IL & RL \\
\midrule
Full \prior{} & 75\% & 100\% & 50\% & 100\% & 45\% & 95\% & 55\% & 100\% \\
w/o low-dimensional bottleneck & 40\% & 35\% & 65\% & 85\% & 15\% & 80\% & 5\% & 20\% \\
w/o history-conditioned encoder & 70\% & 95\% & 35\% & 90\% & 40\% & 60\% & 15\% & 50\% \\
Raw BC (no \prior{}) & 0\% & 15\% & 20\% & 0\% & 0\% & 0\% & 0\% & 0\% \\
\bottomrule
\end{tabular}}
\vspace{0.25em}
\caption{Ablations of \prior{} with a raw-space reference. Each task reports imitation-learning (IL) success and final online residual-RL (RL) success.}
\label{tab:ablations}
\vspace{-0.8em}
\end{table}

\subsection{Action-Flow Visualization}
\label{subsec:failure}

\begin{figure}[H]
    \centering
    \includegraphics[width=\linewidth]{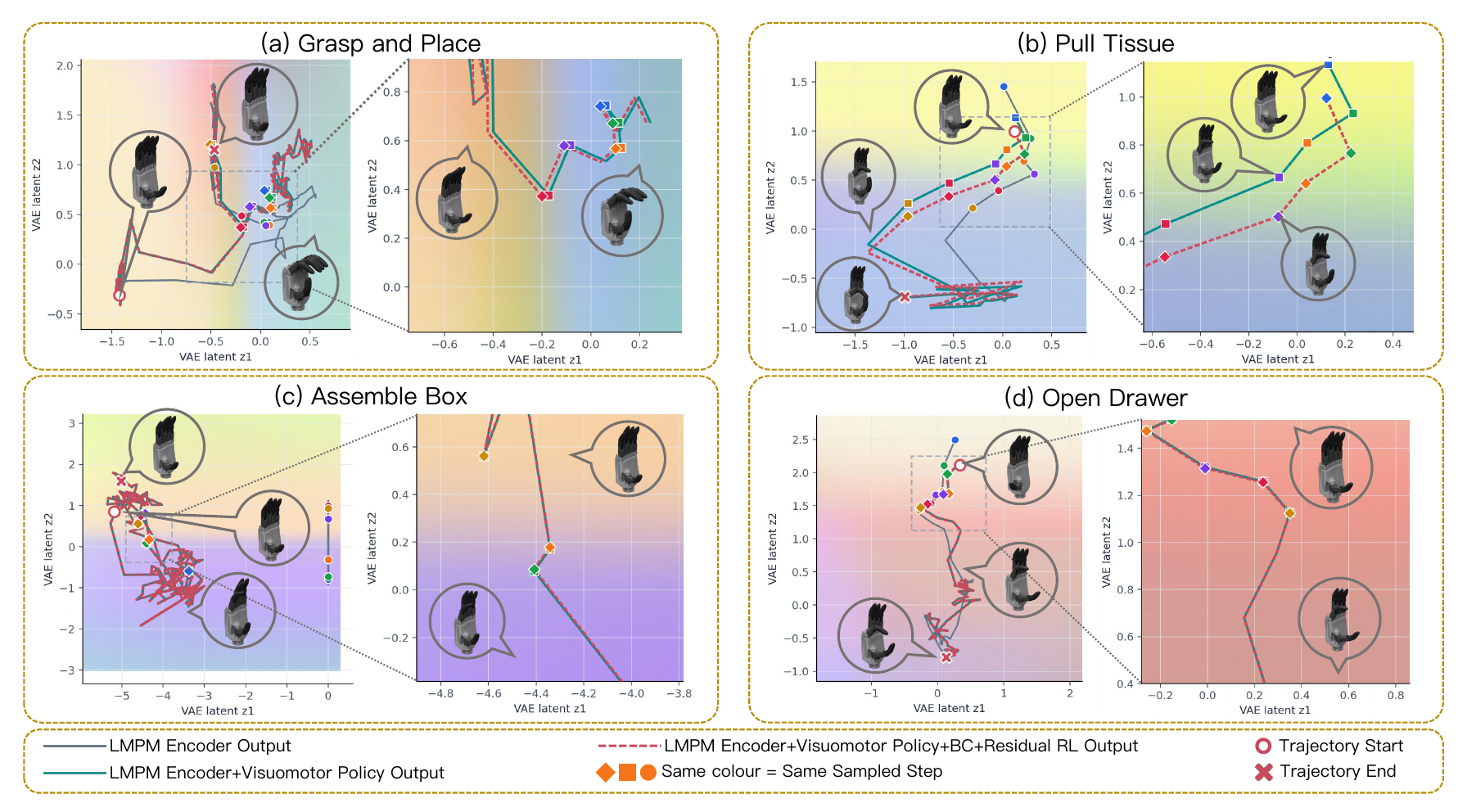}
    \caption{Latent action-flow visualization for the four tasks. Gray trajectories show the \prior{} encoder output, teal trajectories show the latent action after adding the IL-predicted latent offset, and red dashed trajectories show the final latent action after applying the RL latent residual. Hand icons show decoded hand gestures at representative latent locations, and markers with the same color indicate the same sampled timestep.}
    \label{fig:failure}
\end{figure}

Figure~\ref{fig:failure} visualizes the latent action flow during real-world rollouts, showing how actions evolve in the learned hand space across the three stages of \method{}. The history-conditioned prior keeps the hand action on a feasible motion manifold, the IL offset supplies the main vision-guided displacement, and the RL residual makes local online adjustments in the same latent coordinates. When IL leaves a larger gap, as in Grasp \& Place and Pull Tissue, the residual shifts the action toward nearby regions that better complete the task. When IL is already close to success, as in Open Drawer and Assemble Box, the residual remains small and structured. This keeps exploration local while preserving enough freedom to improve contact-rich behaviors.

\section{Limitations}
\label{sec:limitations}

Our current \prior{} is trained from task-specific offline hand-motion data. This makes the latent interface well matched to each task's contact patterns, but deploying \method{} to a new task still requires suitable hand-motion trajectories and prior pretraining. Learning a multi-task or more general hand-motion prior will likely require broader datasets and higher-capacity models, which we leave for future work.

We validate the latent-prior idea on a 6-DoF hand and map the raw hand space to a 2-D latent space. A bimanual manipulation study with 15-DoF Cyberglove measurements found that the first four PCs explain more than 95\% of postural variability~\citep{jarrasse2014analysis}, suggesting that higher-DoF hands may also admit compact hand-motion coordinates. Validating \prior{} on more dexterous hands and broader contact regimes remains future work.

\section{Conclusion}
\label{sec:conclusion}

We presented \method{}, a real-world dexterous learning framework built around \prior{}, a history-conditioned latent motion prior for hand actions. \method{} exposes high-dimensional hand motion through a compact, continuous, and decodable interface shared by the full imitation-to-reinforcement-learning pipeline: behavior cloning predicts offsets around the current prior, and residual reinforcement learning refines the same latent hand action before decoding. This design simplifies high-dimensional hand-action prediction during imitation learning and keeps online exploration close to demonstrated, contact-consistent hand motion. Across four real-world tasks, \method{} outperforms raw, linear, and discrete hand-action representations, supporting latent motion priors as a practical action interface for real-robot dexterous manipulation.

\bibliography{references}

\clearpage
\newcommand{\LAMPMainPaper}{}
\ifdefined\LAMPMainPaper
\else
\documentclass{article}

\usepackage{float}
\usepackage{wrapfig}
\usepackage[preprint]{corl_2026}
\usepackage{amsmath,amssymb,mathtools}
\usepackage{booktabs}
\usepackage{array}
\usepackage{tabularx}
\usepackage{graphicx}
\usepackage{microtype}

\setlength{\abovecaptionskip}{3.5pt}
\setlength{\belowcaptionskip}{0pt}
\setlength{\intextsep}{6pt plus 1pt minus 1pt}
\setlength{\textfloatsep}{10pt plus 1pt minus 2pt}
\setlength{\floatsep}{6pt plus 1pt minus 1pt}

\definecolor{lampred}{rgb}{0.808,0.204,0.255}
\definecolor{lampcoral}{rgb}{0.929,0.443,0.416}
\definecolor{lampyellow}{rgb}{0.992,0.863,0.427}
\definecolor{lampgold}{rgb}{0.745,0.592,0.098}
\newcommand{\lampword}{\textcolor{lampred}{L}\textcolor{lampcoral}{A}\textcolor{lampyellow}{M}\textcolor{lampgold}{P}}

\title{\lampword{}: Latent Motion Prior-Guided Real-World Learning for Dexterous Hand Manipulation}

\author{
\textbf{Xinye Yang}$^{1,3,4,*}$ \quad
\textbf{Zhiyuan Ma}$^{2,3,*}$ \quad
\textbf{Hongze Yu}$^{3,\dagger}$ \quad
\textbf{Yuanpei Chen}$^{3}$ \\
\textbf{Yaodong Yang}$^{5}$ \quad
\textbf{Xiaojie Chai}$^{3}$ \quad
\textbf{Xinlei Chen}$^{2}$ \quad
\textbf{Chao Yu}$^{2,\dagger}$ \\[0.35em]
{\normalfont $^{1}$Fudan University, $^{2}$Tsinghua University, $^{3}$PsiBot} \\
{\normalfont $^{4}$Zhongguancun Academy, $^{5}$Peking University} \\
{\normalfont $^{*}$Equal contribution, $^{\dagger}$Corresponding authors}
}

\newcommand{\method}{LAMP}
\newcommand{\obs}{o}
\newcommand{\act}{a}
\newcommand{\arm}{\mathrm{arm}}
\newcommand{\hand}{\mathrm{hand}}
\newcommand{\hist}{H}
\newcommand{\latent}{z}
\newcommand{\real}{\mathbb{R}}
\newcommand{\Aarm}{\mathcal{A}_{\arm}}
\newcommand{\Ahand}{\mathcal{A}_{\hand}}
\newcommand{\Zspace}{\mathcal{Z}}
\newcommand{\Ddemo}{\mathcal{D}_{\mathrm{demo}}}
\newcommand{\Donline}{\mathcal{D}_{\mathrm{online}}}
\newcommand{\prior}{LMPM}
\newcommand{\figbox}[2][0.16\textheight]{%
  \fbox{\begin{minipage}[c][#1][c]{0.96\linewidth}
  \centering\small #2
  \end{minipage}}}

\hypersetup{
  pdftitle={LAMP: Latent Motion Prior-Guided Real-World Learning for Dexterous Hand Manipulation -- Supplementary Material},
  pdfauthor={Xinye Yang, Zhiyuan Ma, Hongze Yu, Yuanpei Chen, Yaodong Yang, Xiaojie Chai, Xinlei Chen, Chao Yu},
  pdfsubject={arXiv preprint supplementary material},
  pdfkeywords={Dexterous Manipulation, Motion Prior, Real-World Robot Learning}
}

\begin{document}
\maketitle
\begin{center}
{\large Supplementary Material}
\end{center}
\appendix
\numberwithin{equation}{section}
\numberwithin{figure}{section}
\numberwithin{table}{section}
\raggedbottom
\fi

\ifdefined\LAMPMainPaper
\appendix
\numberwithin{equation}{section}
\numberwithin{figure}{section}
\numberwithin{table}{section}
\else
\fi

\section{Off-Manifold Exploration Analysis}
\label{app:manifold_exploration}

Online residual RL should explore near the demonstrated hand-motion manifold: moving too far away from it often breaks contact before the reward signal can guide recovery. We therefore measure how much of a small exploration step leaves this manifold. For each method, we match the decoded hand-action displacement budget, denoted $\mathrm{Disp}$, and compute the increase in nearest-neighbor distance to the demonstrated hand actions, denoted $\Delta\mathrm{NN}$. The ratio $\Delta\mathrm{NN}/\mathrm{Disp}$ measures the fraction of the step that points away from the data-supported motion manifold.

\begin{figure}[h]
    \centering
    \includegraphics[width=\linewidth]{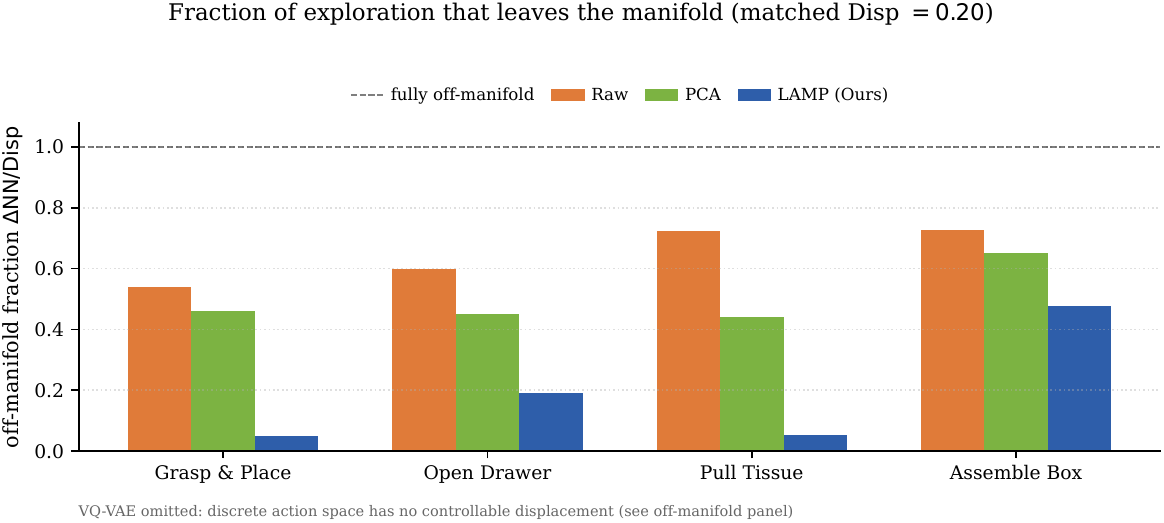}
    \caption{\textbf{Fraction of exploration that leaves the manifold} at a matched displacement budget ($\mathrm{Disp}=0.2$). Bars show $\Delta\mathrm{NN}/\mathrm{Disp}$; $1.0$ (dashed) is fully off-manifold. LAMP keeps the smallest off-manifold fraction on most tasks. VQ-VAE is omitted here because its discrete action space cannot realise a controllable displacement budget; its behaviour is shown in the off-manifold panel instead.}
    \label{fig:app_manifold_ratio}
\end{figure}

This normalization separates the size of an exploration step from its direction. A value near $0$ means that the decoded perturbation mostly moves along nearby demonstrated motions; a value near $1$ means that the same displacement is almost entirely off-manifold. Figure~\ref{fig:app_manifold_ratio} reports the ratio at $\mathrm{Disp}=0.2$ for the continuous action interfaces. Raw exploration is consistently far from the manifold, and PCA remains substantially off-manifold because its fixed linear coordinates do not follow the curved hand-motion structure. LAMP keeps the ratio lowest on most tasks, showing that its continuous latent residuals provide a better local neighborhood for real-robot exploration.

\section{Implementation Details}
\label{app:method_details}

Unless otherwise specified, all action-interface variants use the same demonstrations, camera inputs, arm-action parameterization, behavior-cloning backbone, residual-RL algorithm, and evaluation protocol.

\subsection{Observations and Actions}
\label{app:impl_observation_action}

At each control step, the policy receives the front and wrist RGB images, the current 12-D robot state, and the most recent eight hand targets. The robot action is represented as a 6-D arm command concatenated with a 6-D Ruiyan-hand command. Demonstrations store the hand command as an absolute target. During robot execution, the environment wrapper converts the decoded absolute hand target into the low-level hand delta used by the hardware interface.

Behavior cloning is trained as one-step target prediction. Given the observation at time $t$ and the current action/state $\act_t$, the supervised target is the next 12-D action $\act_{t+1}$. At the start of a trajectory, missing hand-history entries are padded with the first recorded hand target.

\subsection{LMPM Pretraining and Behavior Cloning}
\label{app:impl_lmpm_pretraining}

\prior{} is trained only on the hand-motion component of the demonstrations. Its encoder maps an 8-step hand-target history to a Gaussian latent prior in $\real^2$, and its decoder maps a latent sample to the next 6-D absolute hand target. Both encoder and decoder are MLPs with hidden width 256. We train with a KL weight of $10^{-3}$.

All action interfaces are built from the same small offline set for each task: 50 demonstrations for Grasp \& Place, 20 for Open Drawer, 20 for Pull Tissue, and 30 for Assemble Box. We use a 9:1 split of these demonstrations throughout supervised training: the training split is used to fit \prior{} from hand-motion sequences and to train the BC policy, while the held-out split is used for \prior{} testing and BC validation. All action-interface variants use the same split.

The BC policy uses one frozen ImageNet-pretrained ResNet-18 encoder for each RGB view. Image features are concatenated with the robot state and hand-history features, and the CoreActionHead MLP has hidden sizes $[512,512,256]$. For \method{}, the head predicts a 6-D arm command and a 2-D latent offset. The hand target is decoded as $D_\theta(\mu_t+\Delta z_t)$, where $\mu_t$ is the frozen \prior{} prior mean computed from the hand history. The supervised hand loss is applied after decoding in the 6-D hand-target space.

\begin{table}[H]
\centering
\small
\setlength{\tabcolsep}{4pt}
\renewcommand{\arraystretch}{1.08}
\begin{tabularx}{\linewidth}{@{}>{\raggedright\arraybackslash}p{0.32\linewidth}XX@{}}
\toprule
Setting & \prior{} pretraining & BC policy \\
\midrule
Input history & 8 hand targets & 8 hand targets \\
Output target & next 6-D hand target & next 12-D robot action \\
Network & MLP encoder/decoder, width 256, $d_z=2$ & frozen ResNet-18 per view; MLP head $[512,512,256]$ \\
Optimizer & AdamW & AdamW \\
Batch size & 256 & 128 \\
Learning rate & $2\times10^{-3}$ & $5\times10^{-4}$ \\
Training steps & 20k & 20k \\
Warmup & 500 LR steps; 2000 KL steps & 500 LR steps \\
Noise augmentation & hand-history std 0.01 & arm-state and hand-history std 0.10 \\
Gradient clip & 1.0 & 1.0 \\
\bottomrule
\end{tabularx}
\caption{Training settings for \prior{} and the behavior-cloning policy.}
\label{tab:app_lmpm_bc_hparams}
\end{table}

\subsection{Residual RLPD and Evaluation}
\label{app:impl_residual_rl}

For online improvement, we freeze the BC policy and train a residual SAC/RLPD agent. The residual actor is a tanh-squashed Gaussian policy in the BC core-action space. For \method{}, this space is 8-D: six arm coordinates and two latent hand coordinates. The residual is added to the stopped-gradient BC core action, then the hand component is decoded by the corresponding hand interface in Appendix~\ref{app:impl_bc_interfaces}. The critic is trained on the executed 12-D environment action.

For residual RL, the full offline demonstration set described in Appendix~\ref{app:impl_lmpm_pretraining} is loaded into the demonstration buffer before online interaction begins, while newly collected robot transitions are appended to the online replay buffer. The learner samples half of each batch from online replay and half from the demonstration buffer. With critic-to-actor ratio 2, each learner iteration performs one critic-only update followed by one update of the critic, actor, and entropy temperature. Real-world evaluation uses randomized task starts with fixed policy parameters. Sparse rewards and evaluation success labels are produced by the visual classifier in Appendix~\ref{app:reward_classifier}.

Within each task, all action-interface variants use the same online training budget: 20k steps for Grasp \& Place, 30k for Open Drawer, 40k for Pull Tissue, and 25k for Assemble Box.

\begin{table}[H]
\centering
\small
\setlength{\tabcolsep}{4pt}
\renewcommand{\arraystretch}{1.08}
\begin{tabularx}{\linewidth}{@{}>{\raggedright\arraybackslash}p{0.30\linewidth}X@{}}
\toprule
Setting & Value \\
\midrule
Actor / critic MLP & $[256,256]$, tanh activations, layer normalization \\
Critic ensemble size & 2 \\
Discount & 0.97 \\
Batch size & 256 = 128 online + 128 demonstration \\
Critic-to-actor ratio & 2 \\
Replay capacity & 200k transitions \\
Training starts & 100 online transitions \\
Network publish interval & 50 learner steps \\
Initial temperature & $10^{-2}$ \\
\bottomrule
\end{tabularx}
\caption{Residual-RLPD hyperparameters shared by all action-interface variants.}
\label{tab:app_rl_reward_hparams}
\end{table}

\subsection{Compared Action Interfaces}
\label{app:impl_bc_interfaces}

Only the hand-action interface changes across Raw, PCA, VQ-VAE, and \method{}. The arm command is always predicted in the native 6-D arm-command space. The resulting hand representation is decoded or transformed back to a 6-D absolute hand target before being concatenated with the arm command.

\begin{table}[H]
\centering
\small
\setlength{\tabcolsep}{4pt}
\renewcommand{\arraystretch}{1.08}
\begin{tabularx}{\linewidth}{@{}>{\raggedright\arraybackslash}p{0.15\linewidth}>{\raggedright\arraybackslash}p{0.31\linewidth}X@{}}
\toprule
Interface & Policy hand output & Conversion to 6-D hand target \\
\midrule
Raw & 6-D hand target & identity \\
PCA & 2-D PCA coordinate & inverse PCA transform fitted on demonstration hand targets \\
VQ-VAE & 16-way residual-VQ code & two residual quantizers with four codes each \\
\method{} & 2-D latent offset $\Delta z_t$ & frozen \prior{} decoder $D_\theta(\mu_t+\Delta z_t)$ \\
\bottomrule
\end{tabularx}
\caption{Hand-action interfaces used with the same BC and residual-RL pipeline.}
\label{tab:app_action_interfaces}
\end{table}

\section{Visual Reward Classifier}
\label{app:reward_classifier}

We train a task-specific binary visual classifier for sparse online rewards and evaluation success labels. The classifier takes the processed front and wrist RGB views used by the policy, denoted by the \texttt{global} and \texttt{wrist} image keys, and does not use proprioceptive state. The two view features are concatenated and mapped to a single success logit $f_{\mathrm{cls}}(\obs)$; Table~\ref{tab:app_reward_classifier} gives the architecture and training settings.

Classifier data are collected separately for each task in the real environment. The operator marks successful states during collection, and unmarked visited states from the same sessions are saved as negative examples. Training samples half of each minibatch from the positive replay buffer and half from the negative replay buffer. Let
$p_{\mathrm{cls}}(\obs)=\sigma(f_{\mathrm{cls}}(\obs))$. We optimize
\begin{equation}
\mathcal{L}_{\mathrm{cls}}=
-\mathbb{E}_{(\obs,y)}\left[
y\log p_{\mathrm{cls}}(\obs)+(1-y)\log(1-p_{\mathrm{cls}}(\obs))
\right].
\end{equation}
Training uses the task-specific camera crops defined in the environment config, without stochastic image crop augmentation.

During online interaction, the environment wrapper evaluates the classifier after each robot step and returns
\begin{equation}
r_t=\mathbb{1}\left[p_{\mathrm{cls}}(\obs_{t+1})>0.90\right].
\end{equation}
When $r_t=1$, the wrapper also terminates the episode and sets the success flag in the environment info. This 0.90 threshold is used for online RL rewards and evaluation success labels.

\begin{table}[H]
\centering
\small
\setlength{\tabcolsep}{4pt}
\renewcommand{\arraystretch}{1.08}
\begin{tabularx}{\linewidth}{@{}>{\raggedright\arraybackslash}p{0.28\linewidth}X@{}}
\toprule
Item & Setting \\
\midrule
Input & \texttt{global} and \texttt{wrist} RGB views; no proprioception \\
Visual encoder & One frozen ImageNet-pretrained SERL ResNet-10 per view \\
Pooling & Spatial learned embeddings with 8 blocks and a 256-D bottleneck \\
Classifier head & Dense 256, dropout 0.1, layer norm, ReLU, Dense 1 \\
Training labels & Operator-marked success states as positives; unmarked visited states as negatives \\
Batching & 256 total: 128 positive and 128 negative examples \\
Optimizer & Adam with learning rate $10^{-4}$ \\
Training length & 250 epochs \\
Deployment rule & Sparse reward and success when $p_{\mathrm{cls}}(\obs)>0.90$ \\
\bottomrule
\end{tabularx}
\caption{Visual reward classifier settings used for online RL and evaluation.}
\label{tab:app_reward_classifier}
\end{table}

\section{Teleoperation and Data Collection}
\label{app:data_collection}

\begin{wrapfigure}{r}{0.50\linewidth}
    \vspace{-0.7em}
    \centering
    \includegraphics[width=\linewidth]{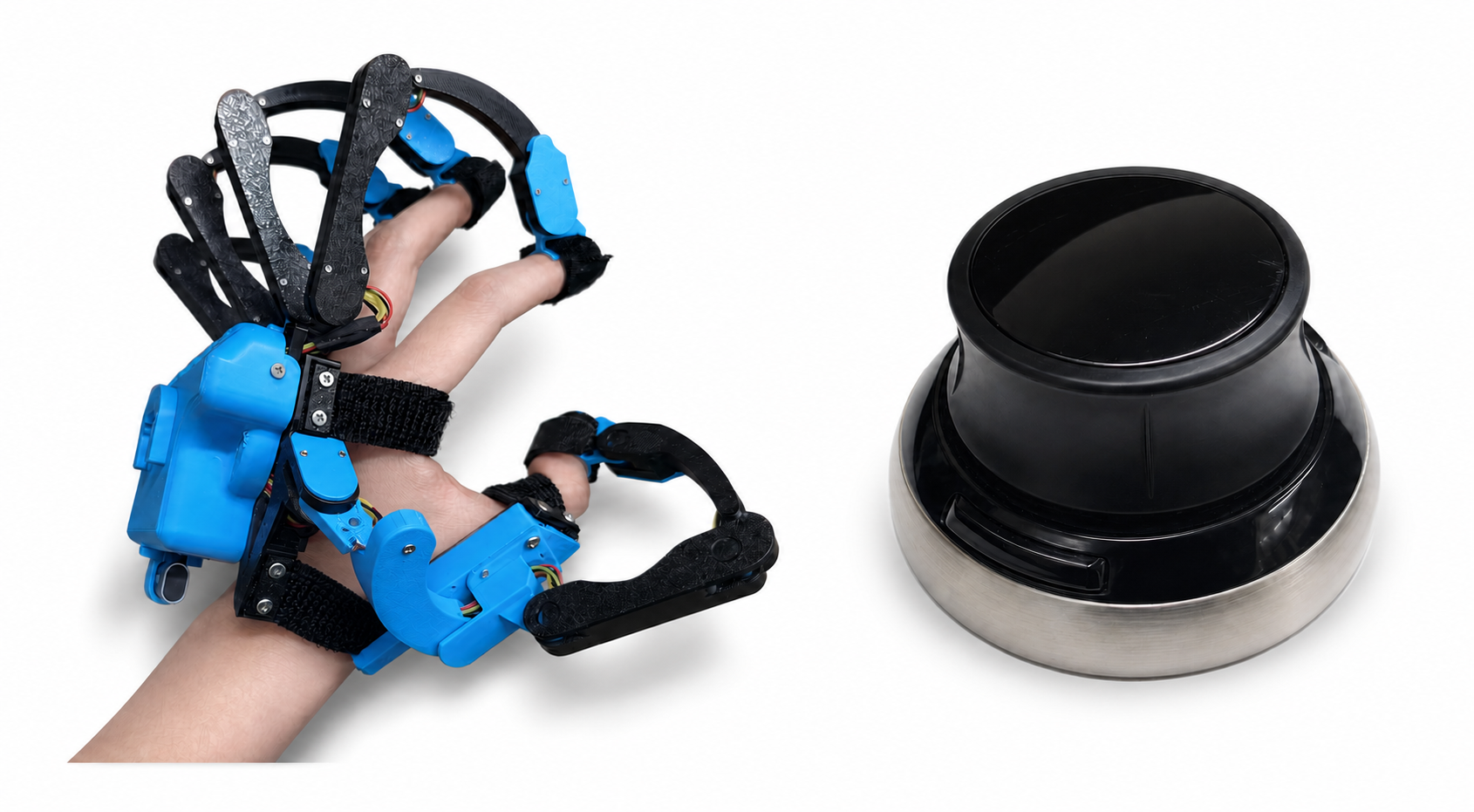}
    \caption{Teleoperation devices. The glove controls the Ruiyan hand; the SpaceMouse controls the end-effector pose.}
    \label{fig:app_data_collection}
    \vspace{-0.8em}
\end{wrapfigure}

Demonstrations are collected with the teleoperation devices in Figure~\ref{fig:app_data_collection}. The SpaceMouse provides the 6-D end-effector command for TCP translation and rotation, while the glove provides the 6-D Ruiyan-hand command. In the real-robot environment, the teleoperation wrapper reads both devices at each control step. Nonzero SpaceMouse motion overrides the arm command. Holding the SpaceMouse right button enables hand teleoperation: the wrapper subtracts the current glove baseline from the measured glove angles, applies the configured hand-joint limits when present, and concatenates the resulting hand delta with the SpaceMouse arm command. When the right button is not held, only the arm part is replaced and the current hand command is left unchanged.

During demonstration recording, the script sends a zero nominal action to the environment and stores the action actually executed by the wrapper from \texttt{info["intervene\_action"]}. Each saved transition contains synchronized RGB observations, proprioceptive state, the executed 12-D arm-hand action, reward and terminal flags, and environment info.

\clearpage
\section{Qualitative Success and Failure Cases}
\label{app:qualitative_cases}

\begin{figure}[H]
    \centering
    \includegraphics[width=0.92\linewidth]{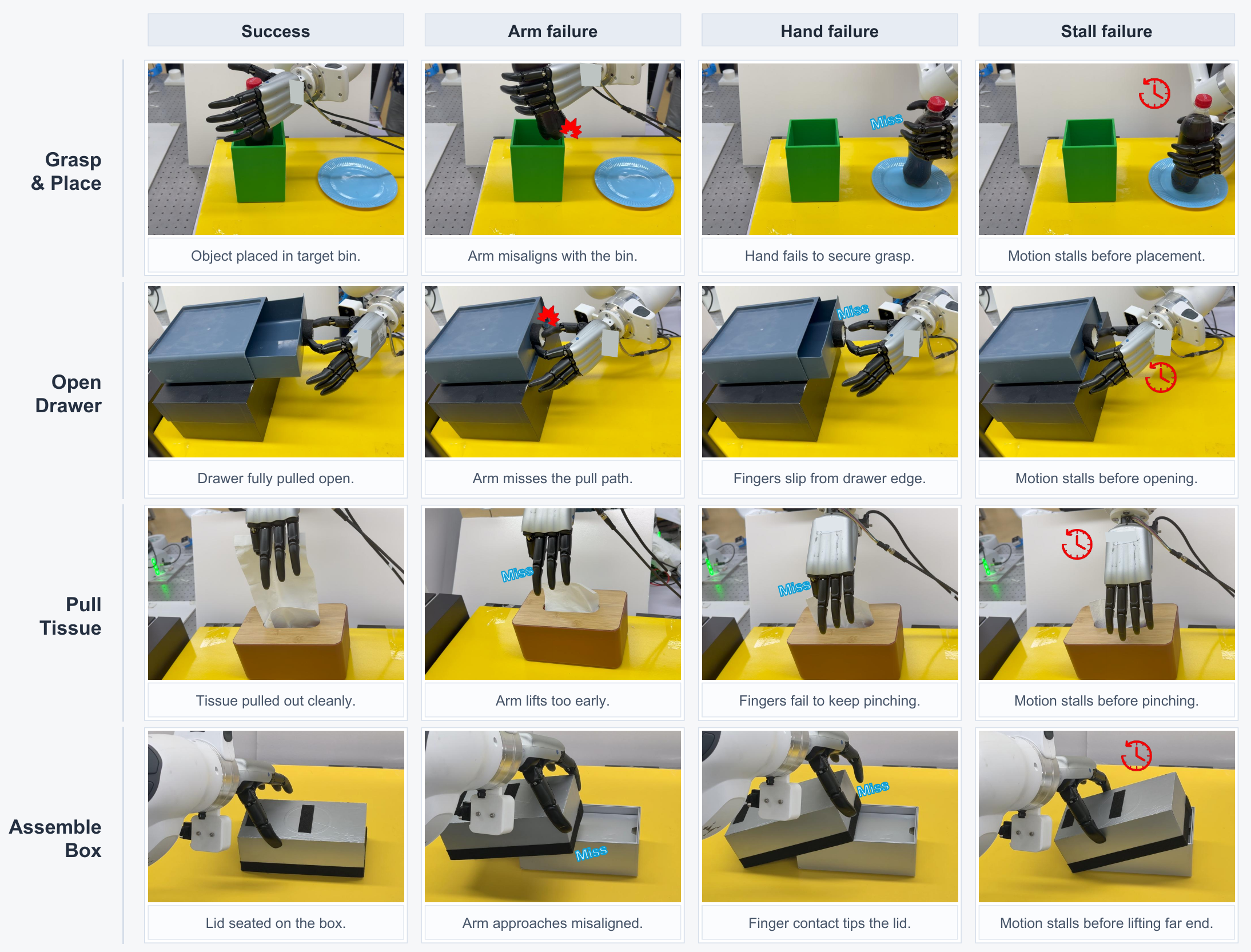}
    \vspace{-0.5em}
    \caption{Qualitative success and failure examples from real-world evaluation rollouts. Rows correspond to tasks; columns show a successful rollout and representative arm, hand, and stall failures.}
    \label{fig:app_qualitative_cases}
    \vspace{-0.8em}
\end{figure}

Figure~\ref{fig:app_qualitative_cases} shows representative evaluation frames for the four real-world tasks. For each task, we compare a successful execution with three recurring failure modes: arm-action errors, dexterous-hand errors, and stalled executions. Each row uses a fixed crop window, so the comparison focuses on the robot-object state rather than camera framing.

We assign a failure to the component that prevents continued task progress in the rollout. Arm failures occur when the end-effector approaches with an unsuitable pose or path, such as missing the drawer pull direction, approaching the tissue off center, or reaching the bin or lid from a pose that cannot complete the interaction. Hand failures occur after the arm reaches the relevant region but the fingers do not maintain the needed contact: the object slips, the drawer edge is not secured, the tissue is pressed instead of pinched, or the lid is tipped away. Stall failures are rollouts that remain near the object but stop making progress before satisfying the task condition.

These examples complement the quantitative failure counts in the main results. Successful executions require both a useful arm trajectory and stable finger contact. Once contact is lost or the object is perturbed into an unfavorable pose, recovery is difficult within the same rollout, which is why contact-preserving hand exploration is important for online refinement.

\begin{figure}[H]
    \centering
    \includegraphics[width=0.96\linewidth]{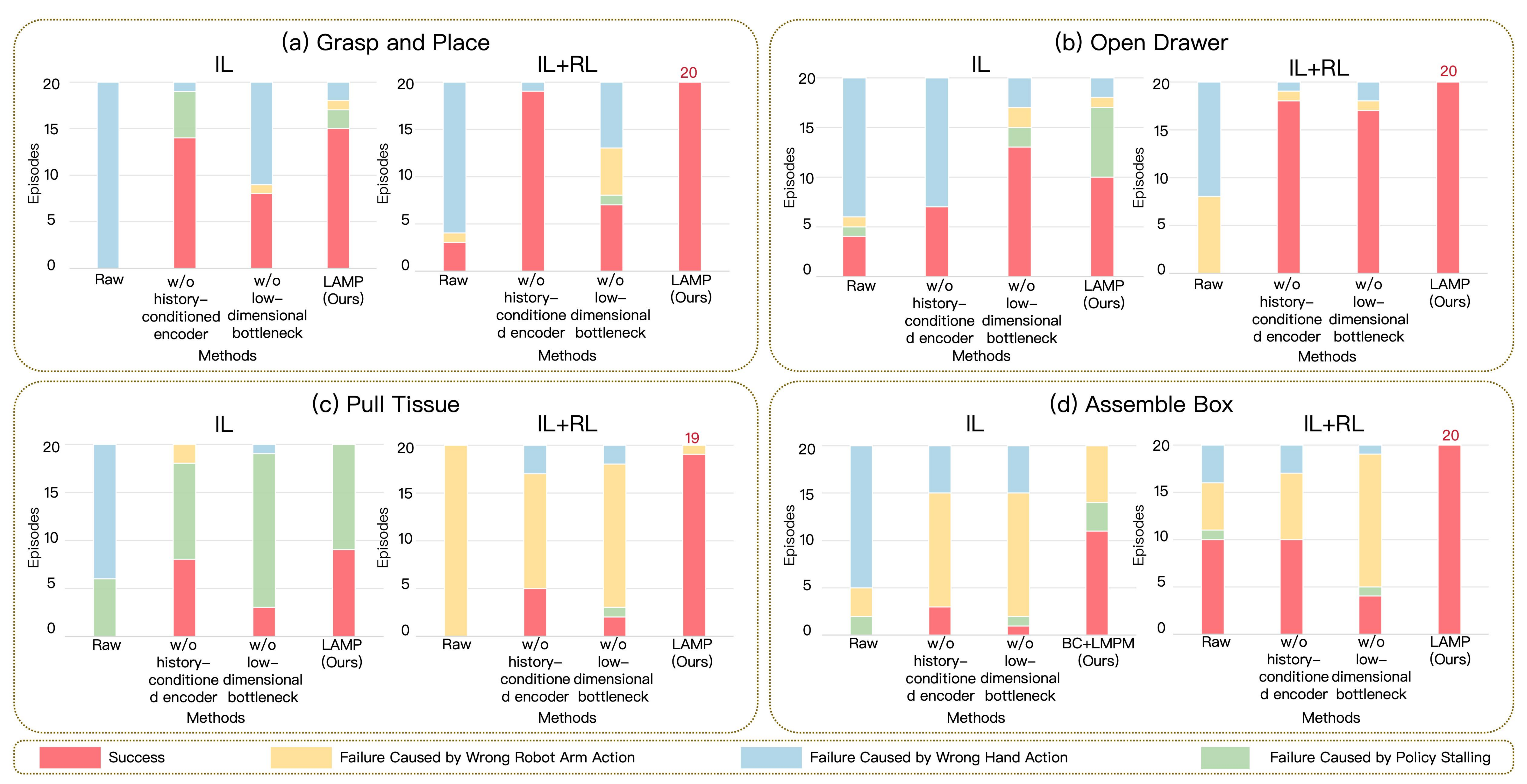}
    \vspace{-0.4em}
    \caption{Ablation outcome breakdown on four real-world tasks. In each task panel, the two subplots correspond to imitation learning (IL) and the final policy after residual RL (IL+RL); the x-axis lists methods and the y-axis reports evaluation episodes. Each stacked bar partitions the 20 evaluation episodes into successes and failures caused by wrong arm action, wrong hand action, or policy stalling. Raw directly regresses the original hand action, w/o history-conditioned encoder removes the prior center computed from recent hand motion, and w/o low-dimensional bottleneck removes the 2-D latent bottleneck while keeping the remaining pipeline unchanged.}
    \label{fig:app_ablation_breakdown}
    \vspace{-0.8em}
\end{figure}

Figure~\ref{fig:app_ablation_breakdown} further breaks down the main ablation results by failure category. The low-dimensional bottleneck helps arm-hand coordination rather than only improving the fingers in isolation. After residual RL, removing the bottleneck leaves many arm-action failures, most visibly on Pull Tissue and Assemble Box, whereas the full \prior{} model has only one arm-action failure across the four tasks. Since the visuomotor policy predicts arm and hand commands jointly, a less compact hand representation also changes the learning problem faced by the arm branch: the policy must fit and explore a larger coupled action space, and the end-effector is more often driven into poses from which the remaining hand motion cannot recover the task.

The history-conditioned encoder has a less category-specific but still visible role. Without the encoder, the policy no longer receives a local prior center from recent hand motion and must predict latent hand commands without this phase reference. The decoder still keeps actions on a learned hand-motion manifold, so residual RL can recover well on the shorter Grasp and Place and Open Drawer tasks. On Pull Tissue and Assemble Box, however, the same ablation leaves mixed arm and hand failures after RL, consistent with poorer timing between the arm approach and the hand contact state. In contrast, the full \prior{} interface provides both a compact bottleneck and a history-conditioned local coordinate system, making post-training failures sparse rather than shifting them between components.

\section{Action Smoothness Analysis}
\label{app:smoothness}

In contact-rich dexterous manipulation, hand-command jitter can break established contacts, induce slips or drops, and make real-robot exploration less recoverable. We analyze the hand-action histories saved during IL and RL evaluation rollouts. To separate high-frequency jitter from deliberate hand opening or closing, we use the second-order hand-target variation
\begin{equation}
J=\frac{1}{T-2}\sum_{t=1}^{T-2}\|\act_{t+2}^{\hand}-2\act_{t+1}^{\hand}+\act_t^{\hand}\|_2 ,
\end{equation}
where $\act_t^{\hand}$ is the executed 6-D absolute hand target at time step $t$. Lower values indicate smoother hand motion, while large values correspond to rapid changes in the direction or magnitude of the hand command.

\begin{figure}[H]
    \centering
    \includegraphics[width=\linewidth]{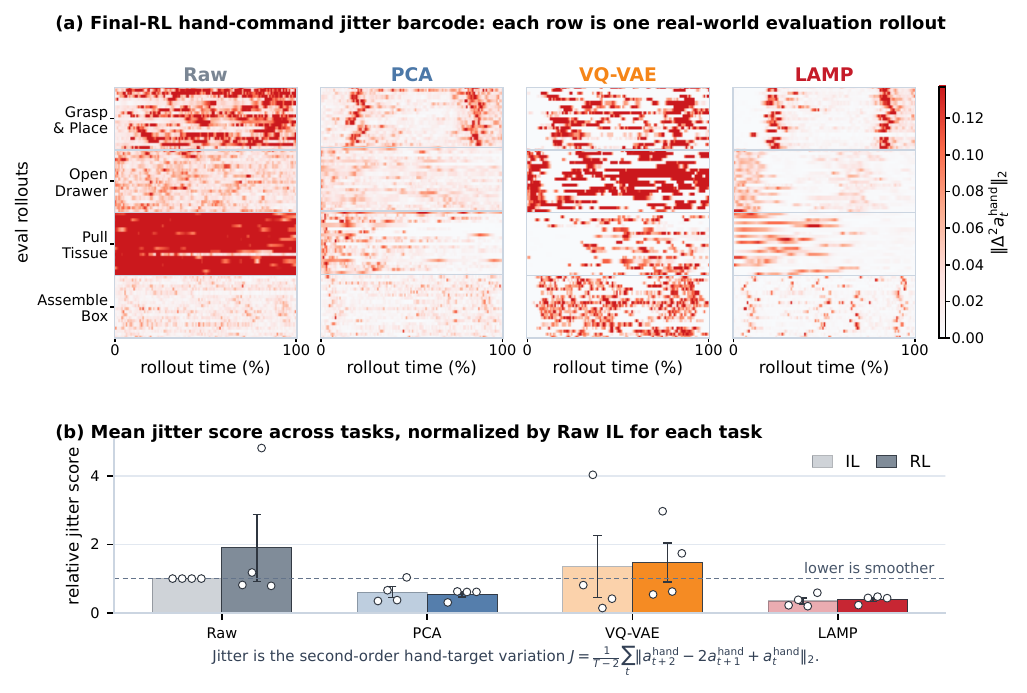}
    \caption{Hand-command jitter during real-world evaluation. Top: final-RL jitter barcode. Within each method panel, the vertical axis stacks individual evaluation rollouts grouped by task; each horizontal row is one rollout, not a hand dimension. The horizontal axis is normalized rollout time, and color denotes the per-time-step term $\|\Delta^2 \act_t^{\hand}\|_2=\|\act_{t+2}^{\hand}-2\act_{t+1}^{\hand}+\act_t^{\hand}\|_2$ used in Eq.~(F.1). The rollout-level jitter score $J$ is the average of these colored values over time. Bottom: task-normalized jitter scores for IL and RL evaluation rollouts; each task is normalized by its Raw-IL score, so lower values indicate smoother hand commands.}
    \label{fig:app_smoothness}
\end{figure}

Figure~\ref{fig:app_smoothness} visualizes the same metric in two ways. The top row shows final-RL evaluation rollouts as a jitter barcode: each row is one rollout and red bands indicate high second-order command variation. Raw and VQ-VAE policies exhibit frequent high-jitter bands, while \method{} keeps most rollouts close to the low-jitter range. The bottom plot aggregates IL and RL rollouts across tasks after normalizing each task by its Raw-IL jitter score.

\ifdefined\LAMPMainPaper
\else
\end{document}
\fi

\end{document}